\documentclass[twoside,11pt]{article}

\usepackage{jmlr2e}
\usepackage{amsmath}
\usepackage{color}
\usepackage{listings}
\usepackage[autostyle]{csquotes}
\usepackage[colorinlistoftodos]{todonotes}
\usepackage{caption}
\usepackage{graphicx}
\usepackage{subcaption}
\usepackage{wrapfig}
\usepackage[hidelinks]{hyperref}
\usepackage{xspace}
\usepackage{todonotes}

\newcommand{\ffm}{\textit{fast}FM\xspace}

\jmlrheading{17}{2016}{1-5}{7/15; Revised 5/16}{10/16}{Immanuel Bayer}

\ShortHeadings{\ffm: A Library for Factorization Machines}{Bayer}
\firstpageno{1}

\begin{document}

\title{ \ffm: A Library for Factorization Machines}

\author{\name Immanuel Bayer \email immanuel.bayer@uni-konstanz.de \\
       \addr University of Konstanz\\
       78457 Konstanz , Germany}

\editor{Cheng Soon Ong}

\maketitle

\begin{abstract}%   <- trailing '%' for backward compatibility of .sty file
%State the problem
Factorization Machines (FM) are currently only used in a narrow range of applications and
are not yet part of the standard machine learning toolbox, despite their great success
in collaborative filtering and click-through rate prediction.
%Say why it's an interesting problem
%This is a pity, because even though FMs are recognized as being very successful for recommender
%system type applications they are a general model to deal with sparse and high dimensional features.
However, Factorization Machines are a general model to deal with sparse and high dimensional features.
%Say what your solution achieves
Our Factorization Machine implementation (\ffm) provides easy access to many solvers and
supports regression, classification and ranking tasks.
%Say what follows from your solution
Such an implementation simplifies the use of FM for a wide range of applications.
Therefore, our implementation has the potential to improve understanding of the FM model and drive
new development.
\end{abstract}


\section{Introduction}
% Motivate that FM are useful
This work aims to facilitate research for matrix factorization based machine learning (ML)
models. Factorization Machines are able to express many different latent factor models and
are widely used for collaborative filtering tasks \citep{rendle:tist2012}.
An important advantage of FM is that the model equation
$$w_0 \in \mathbb{R}, x, w \in \mathbb{R}^p, v_i \in \mathbb{R}^k$$
\begin{equation}
    \hat{y}^{FM}(x) := w_0 + \sum^p_{i=1} w_i x_i + \sum^p_{i=1} \sum^p_{j>i}
    \langle v_i, v_j \rangle x_i x_j
    \label{eq:fm}
\end{equation}
conforms to the standard notation for vector based ML. FM learn a factorized coefficient
$\langle v_i, v_j \rangle$ for each feature pair $x_ix_j$ (eq. \ref{eq:fm}). This
makes it possible to model very sparse feature interactions,
as for example, encoding a sample as $x = \{\cdots, 0, \overbrace{1}^{x_i},0 , \cdots, 0,  \overbrace{1}^{x_j}, 0, \cdots \}$
yields $\hat{y}^{FM}(x) = w_0 + w_i + w_j + v_i^Tv_j$ which is equivalent to
(biased) matrix factorization $R_{i,j} \approx b_0 + b_i + b_j+ u_i^Tv_j$  \citep{srebro2004maximum}.
Please refer to \citet{rendle:tist2012} for more encoding examples.
FM have been the top performing model in various machine learning competitions \citep{rendle2009factor, rendle2012social, bayer2013factor} with different objectives
(e.g. ‘What Do You Know?’ Challenge\footnote{\url{http://www.kaggle.com/c/WhatDoYouKnow}},
EMI Music Hackathon\footnote{\url{http://www.kaggle.com/c/MusicHackathon}}).
\ffm includes solvers for regression, classification and ranking problems (see Table \ref{tab:solver})
and addresses the following needs of the research community:
(i) easy interfacing for dynamic and interactive languages such as R, Python and
Matlab; (ii) a Python interface allowing interactive work; (iii) a publicly
available test suite strongly simplifying modifications or adding of new features;
(iv) code is released under the \textbf{BSD-license} allowing the integration in (almost)
any open source project.

\section{Design Overview}
\begin{wrapfigure}{l}{40mm}
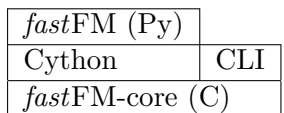

%\begin{table}[h]
    \centering
    \begin{tabular}{|l|l|}
        \cline{1-1}
        \multicolumn{1}{|l|}{\ffm (Py)} \\ \cline{1-2}
        Cython& CLI           \\ \hline
        \multicolumn{2}{|l|}{\ffm-core (C)} \\ \hline
    \end{tabular}
    \caption{Library Architecture}
    \label{tab:architecture}
\end{wrapfigure}

The \ffm  library has a multi layered software architecture (Figure \ref{tab:architecture})
that separates the interface code from the performance critical parts
(\ffm-core). The core contains the solvers, is written in C and can
be used stand alone. Two user interfaces are available: a command line interface (CLI)
and a Python interface. Cython \citep{behnel2011cython} is used to create a Python extension from the C library.
Both, the Python and C interface,
serve as reference implementation for bindings to additional languages.

%\begin{figure*}[h!]
%\end{figure*}
%\end{table}

\subsection{fastFM-core}

FM are usually applied to very sparse design matrices, often with a sparsity over 95 \%,
due to their ability to model interaction between very high dimensional categorical features.
We use the standard compressed row storage (CRS) matrix format as underlying data structure and rely on
the CXSparse\footnote{CXSparse is LGPL licensed.} library \citep{davis2006direct} for fast sparse matrix / vector operations.
This simplifies the code and makes memory sharing between Python and C straight forward.
\\ \indent
\ffm contains a test suite that is run on each commit to the GitHub repository via a continuous integration server\footnote{\url{https://travis-ci.org/ibayer/fastFM-core}}.
Solvers are tested using state of the art techniques, such as Posterior Quantiles \citep{cook2006validation}
for the MCMC sampler and Finite Differences for the SGD based solvers.

\subsection{Solver and Loss Functions}
\ffm provides a range of solvers for all supported tasks (Table \ref{tab:solver}).
The MCMC solver implements the Bayesian Factorization Machine model
\citep{Freudenthaler_bayesianfactorization} via Gibbs sampling. We use
the pairwise Bayesian Personalized Ranking (BPR) loss \citep{Rendle_BBP} for ranking.
More details on the classification and regression solvers can be found in
\cite{rendle:tist2012}.

\begin{table}[h!]
    \centering
    \begin{tabular}{| l | l | l |}
    \hline
    \textbf{Task} &\textbf{Solver} & \textbf{Loss} \\ \hline
    Regression & ALS, MCMC, SGD & Square Loss \\
    \hline
    Classification & ALS, MCMC, SGD &  Probit (MAP), Probit, Sigmoid \\
    \hline
    Ranking & SGD &  BPR \citep{Rendle_BBP}\\
    \hline
    \end{tabular}
    \caption{Supported solvers and tasks}
    \label{tab:solver}
\end{table}

\subsection{Python Interface}
The Python interface is compatible with the API of the widely-used {\tt scikit-learn}
library \citep{scikit-learn} which opens the library to a large user base. The following
code snippet shows how to use MCMC sampling for an FM classifier and how to make predictions on new data.

\begin{lstlisting}[language=Python, frame=single, columns=fullflexible]
fm = mcmc.FMClassification(init_std=0.01, rank=8)
y_pred = fm.fit_predict(X_train, y_train, X_test)
\end{lstlisting}
\ffm provides additional features such as warm starting a solver from a previous solution (see MCMC example).
\begin{lstlisting}[language=Python, frame=single, columns=fullflexible]
fm = als.FMRegression(init_std=0.01, rank=8, l2_reg=2)
fm.fit(X_train, y_train)
\end{lstlisting}

\section{Experiments}
\begin{figure*}[h!]
    \centering
    \includegraphics[width=\textwidth]{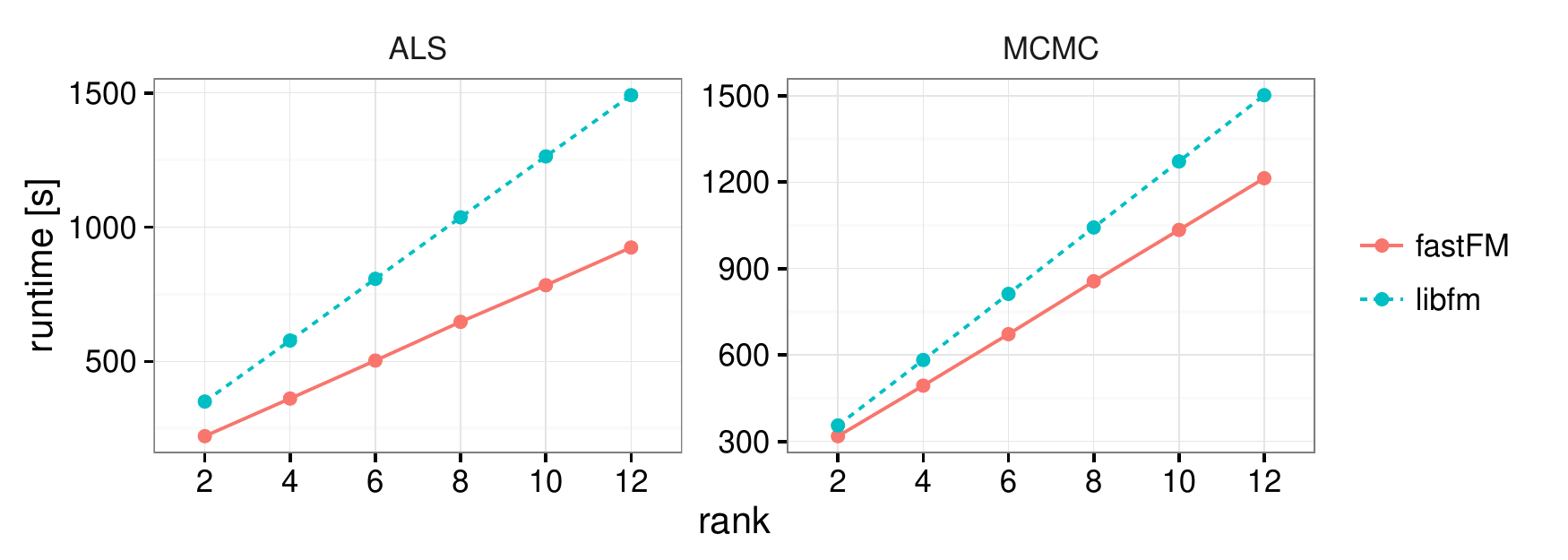}
    \caption{A runtime comparison between \ffm
    and libFM is shown. The evaluation is done on the MovieLens 10M data set.}
 \label{fig:comparision}
\end{figure*}

\begin{figure*}[h]
    \centering
    \begin{subfigure}[b]{0.48\textwidth}
    \includegraphics[width=\textwidth]{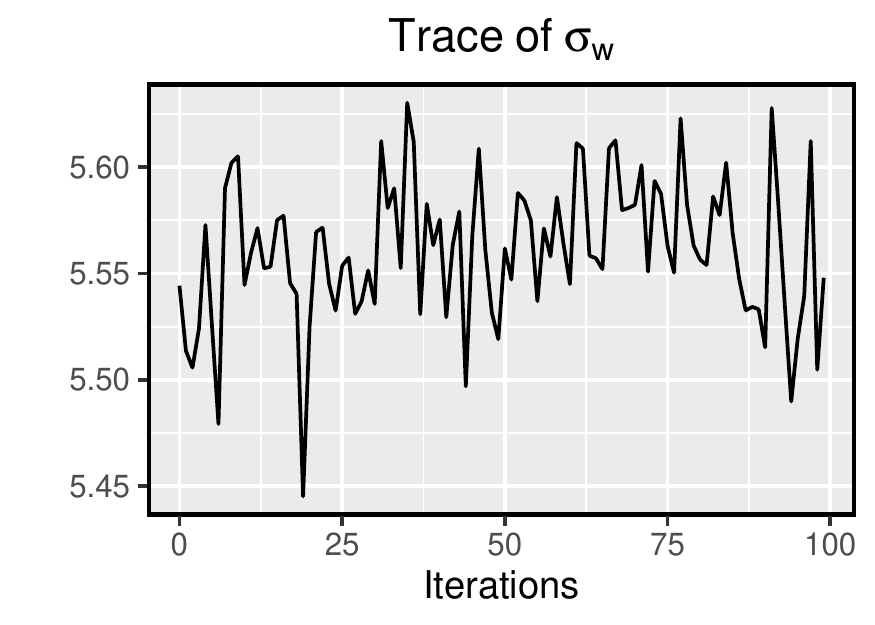}
        %\caption{A gull}
        %\label{fig:gull}
    \end{subfigure}
    ~ %add desired spacing between images, e. g. ~, \quad, \qquad, \hfill etc. 
    \begin{subfigure}[b]{0.48\textwidth}
    \includegraphics[width=\textwidth]{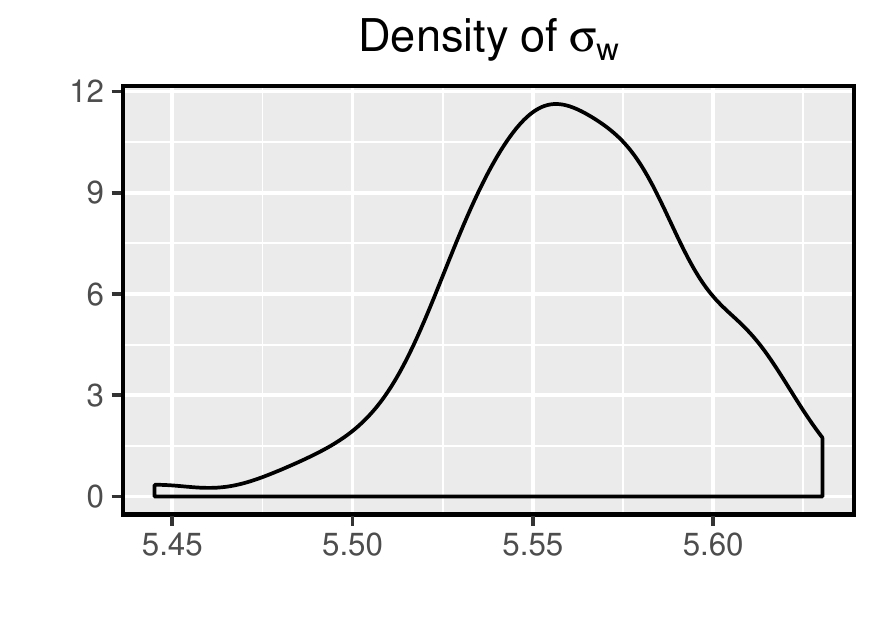}
        %\caption{A tiger}
        %\label{fig:tiger}
    \end{subfigure}
    \caption{MCMC chain analysis and convergence diagnostics example for the
    hyperparameter $\sigma_w$ evaluated on the MovieLens 10M data set.}
    \label{fig:mcmc_trace}
\end{figure*}

libFM\footnote{\url{http://libfm.org}} is the reference implementation for FM and the only one
that provides ALS and MCMC solver.
Our experiments show, that the ALS and MCMC solver in \ffm compare favorable to libFM
with respect to runtime (Figure \ref{fig:comparision}) and are indistinguishable
in terms of accuracy. The experiments have been conducted on the MovieLens 10M data set using the
original split with a fixed number of 200 iterations for all experiments. The x-axis indicates
the number of latent factors (rank), and the y-axis the runtime in seconds. The plots
show that the runtime scales linearly with the rank for both implementations.
The code snippet below shows how simple it is to write Python code that allows model
inspection after every iteration. The induced Python function call overhead occurs only once
per iteration and is therefore neglectable.
This feature can be used for Bayesian Model Checking as demonstrated in Figure \ref{fig:mcmc_trace}.
The figure shows MCMC summary statistics for the first order hyper parameter $\sigma_w$.
Please note that the MCMC solver uses Gaussian priors for the model parameter \citep{Freudenthaler_bayesianfactorization}.

\begin{lstlisting}[language=Python, frame=single, columns=fullflexible]
    fm = mcmc.FMRegression(n_iter=0)
    # initialize coefficients
    fm.fit_predict(X_train, y_train, X_test)

    for i in range(number_of_iterations):
        y_pred = fm.fit_predict(X_train, y_train, X_test, n_more_iter=1)
        # save, or modify (hyper) parameter
        print(fm.w_, fm.V_, fm.hyper_param_)
\end{lstlisting}

Many other analyses and experiments can be realized with a few
lines of Python code without the need to read or recompile the performance critical
C code.

\section{Related Work}
Factorization Machines are available in the large scale machine learning libraries
GraphLab \citep{low2014graphlab} and Bidmach \citep{canny2013bidmach}. The toolkit
Svdfeatures by \cite{chen2012svdfeature} provides a general MF model that is similar to a FM. The implementations
in GraphLab, Bidmach and Svdfeatures only support SGD solvers and don't provide a ranking loss.
It's not our objective to replace these distributed machine learning frameworks:
but to be provide a FM implementation that is easy to use and easy to extend without sacrificing performance.

% Acknowledgements should go at the end, before appendices and references

\acks{This work was supported by the DFG under grant Re 3311/2-1.}

%\newpage
%\clearpage

\appendix
%\listoftodos
%\bibliographystyle{plainnat}
%\bibliographystyle{natbib}
\bibliography{}

\end{document}